# Improved Fitness Dependent Optimizer for Solving Economic Load Dispatch Problem

Barzan Hussein Tahir [1] · Tarik A. Rashid [2] · Hafiz Tayyab Rauf[3] · Nebojsa Bacanin[4]· Amit Chhabra[5] · S. Vimal[6] ·Zaher Mundher Yaseen[7,8,9]∗

[1] Department of Computer Science and Engineering, University of Kurdistan Helwer, Erbil, Iraq; E-mail: barzan.husseintahir@ukh.edu.krd

[2] Department of Computer Science and Engineering, University of Kurdistan Helwer, Erbil, Iraq: E-mail: tarik.ahmed@ukh.edu.krd

[3] Centre for Smart Systems, AI and Cybersecurity, Staffordshire University, Stoke-on-Trent, UK3; E-mail: hafiztayyabrauf093@gmail.com

[4] Singidunum University, Danijelova 32, Belgrade, 11000, Serbia; E-mail: nbacanin@singidunum.ac.rs

[5] Department of Computer Engineering and Technology, Guru Nanak Dev University, Amritsar-INDIA; E-mail: amit.cse@gndu.ac.in

[6] Department of Artificial Intelligence and Data Science, Ramco Institute of Technology, North Venganallur Village, Rajapalayam - 626 117, Virudhunagar District, Tamilnadu, India; E-mail: vimal@ritrjpm.ac.in

[7] Department of Earth Sciences and Environment, Faculty of Science and Technology, Universiti Kebangsaan Malaysia, Bangi 43600, Selangor, Malaysia

[8] Adjunct Research Fellow, USQ's Advanced Data Analytics Research Group, School of Mathematics Physics and Computing, University of Southern Queensland, QLD 4350, Australia

[9] New era and Development in Civil Engineering Research Group, Scientific Research Center, Al-Ayen University, Thi-Qar 64001, Iraq; E-mail: yaseen@alayen.edu.iq (corresponding author)

**Abstract**

Economic Load Dispatch depicts a fundamental role in the operation of power systems, as it decreases the environmental load, minimizes the operating cost, and preserves energy resources. The optimal solution to Economic Load Dispatch problems and various constraints can be obtained by evolving several evolutionary and swarm-based algorithms. The major drawback to swarm-based algorithms is premature convergence towards an optimal solution. Fitness Dependent Optimizer is a novel optimization algorithm stimulated by the decision-making and reproductive process of bee swarming. Fitness Dependent Optimizer (FDO) examines the search spaces based on the searching approach of Particle Swarm Optimization. To calculate the pace, the fitness function is utilized to generate weights that direct the search agents in the phases of exploitation and exploration. In this research, the authors have carried out Fitness Dependent Optimizer to solve the Economic Load Dispatch problem by reducing fuel cost, emission allocation, and transmission loss. Moreover, the authors have enhanced a novel variant of Fitness Dependent Optimizer, which incorporates novel population initialization techniques and dynamically employed sine maps to select the weight factor for Fitness Dependent Optimizer. The enhanced population initialization approach incorporates a quasi-random Sabol sequence to generate the initial solution in the multi-dimensional search space. A standard 24-unit system is employed for experimental evaluation with different power demands. Empirical results obtained using the enhanced variant of the Fitness Dependent Optimizer demonstrate superior performance in terms of low transmission loss, low fuel cost, and low emission allocation compared to the conventional Fitness Dependent Optimizer. The experimental study obtained 7.94E-12, the lowest transmission loss using the enhanced Fitness Dependent Optimizer. Correspondingly, various standard estimations are used to prove the stability of Fitness Dependent Optimizer in phases of exploitation and exploration.

**Keywords**: FDO · Chaotic Maps · Premature Convergence. Optimization Algorithms. Fitness Dependent Optimizer

## 1. Introduction

After the development of computers, the main objective was to investigate unknown solutions and find the best possible solution. During World War II, Alan Turing broke a cipher of Germany named Enigma by using an algorithm used for searching. [1]. Many challenges arose in solving problems in real life due to the improvements in working methods and the exciting acceleration in the extent of computations. Hence, techniques based on numerical programming and conventional logic emerged to overcome the drawbacks of instantly and capably resolving complicated problems [2]. Various algorithms like optimization problems have been designed to manage these limitations. The best possible solution was gained through the optimization method by studying its parameter. All the possible values of present solutions were expressed as a set one of which is the fittest solution. Usually, problems of optimization are solved to design algorithms of optimization [3]. Optimization algorithms are classified into two groups i.e., stochastic algorithms and deterministic algorithms [4], [5]. Deterministic algorithms generate a group of related answers when the iterations are started by using an introductory initial point all this happened by using inclination [6]. On the other hand, stochastic algorithms constantly generate distinct answers with related values in the absence of inclination. Diversely, concluding values have a slight difference. There are two categories of stochastic algorithms i.e. metaheuristic and heuristic [7], [8].

Heuristic algorithms use the trial-and-error method to find a solution, and it is supposed that these algorithms will consume reasonable time to reach a solution [9], [10]. Moreover, heuristic algorithms are aimed to utilize various methods in local examinations and techniques of randomization [11]. Further analysis and advancements were made in heuristic algorithms and converted to metaheuristic algorithms [12], [13]. The novel collections of algorithms have better performance than heuristic algorithms; accordingly, the affix of 'meta' that means 'far off' was linked with these algorithms.

Recently, available problems of the real world have turned complicated in considerations of cost, time, and space; it is not possible to traverse all credible solutions. Hence, fast and low-cost techniques are required [14], [15]. Thus, scientists studied the natural events and behaviors of animals to resolve these issues, like how ants select their paths, how fish, flies, or birds chase their prey, and the working of gravity. So, all the algorithms inspired by nature are called nature-inspired algorithms [16]. FDO algorithm that is also known as fitness-dependent optimizer was introduced by Jaza Abdullah and Tarik Rashid in 2019. FDO algorithm studies the bee swarms' reproduction practices and follows the activities of swarms. This algorithm finds out the best solution among the pool of solutions [17]. Intelligent computations are rising in various areas of research because of their capability to integrate with large complex and interconnected systems with high speed and accuracy.

Economic load dispatch (ELD) is the most vital and significant field of power system planning and operation [18], [19]. The chief aim of ELD was to list a group of real power provided by resources of online generation to satisfy the lacked demand whenever needed under a group of limitations [20], regarding system and unit technical constraints with the least production cost. In general, the ELD problem can be validated as an extremely non-linear and non-smooth stifled problem of optimization usually for huge systems. The cost of fuel is concerned with the varying costs of the generation of electricity.

The main purpose of addressing the ELD problem is to determine the necessary output power to satisfy the system's requirements in such a way that the cost is limited to its possible minimum value and limits, such as Prohibited Operating Zone (POZ) and Valve-Point Effects (VPE) [20]. There is a need for efficient ways of producing electricity. Increasing the cost of fuel makes the method of power production expensive. Advanced systems are therefore expected to propose an economical generation, delivery, and transmission method while keeping electrical limitations in mind. The total device requirement is divided into several units by ELD, which reduces the total cost of generation. Several complications exist while obtaining the ELD problem's global best solution; results are less accurate due to the non-linear nature of classical methods, confined to convergence issues, and best local solutions [6].

The major drawback of some techniques like evolutionary computing is premature convergence [21]. In heuristics, exploitation, and exploration perform a significant role. The capacity of an algorithm to hunt globally is called exploration and the ability to search locally is known as exploitation [22]. The stability of exploration and exploitation highly affects the swarm-based algorithms' performance. Smaller exploration and extreme exploitation lead to premature convergence, on the other hand, more limited exploitation and higher exploration can cause barriers to gaining the best solution [23].

In this research, by reducing fuel expense, emission allocation, and transmission loss, the authors have implemented FDO to solve the ELD problem. Similarly, authors have enhanced a new FDO variant that consolidates specific techniques of population initialization and manipulates sinus maps to pick the FDO weight factor dynamically. The enhanced population initialization method combines a quasi-random Sabol sequence to generate the initial solution for the multi-dimensional search space. A typical 24-unit device is implemented with different power demands for preliminary evaluation. The observational results obtained by implementing the Enhanced FDO variant illustrate the outstanding performance compared to the standard FDO in low transmission loss, low fuel cost, and low emission allocation.

## 2 Related Work

In this section, various studies have been reported similar to the enhanced approach but with different evolutionary and swarm-based algorithms i.e PSO, BA, DE, GA, BCO, and other combinations of the recent state-of-the-art algorithms.

Authors carried out a study to suggest a method for the solution of ELD problems and to contrast it with various available solutions [24]. Their enhanced method owns the following characteristics, managing issues of non-differentiability, all the problems caused by PBC are also resolved, and problems of multi-objective nature of PBC are also resolved. They implemented their approach on 5 generation systems, the achieved outcomes proved that more effective Pareto-curve is gained. An enhanced algorithm inspired by nature i.e Bat algorithm that offers firm convergence and excellent computational performance was conducted by [7]. Yang enhanced the BAT algorithm in 2010 after inspiring by bat's echolocation behavior. They stated that this bat quality enables the bat to locate the prey, i.e., various insects, even in the absence of light. Their enhanced method aimed to reduce the total cost of generation in the case of a thermal power plant.

In another research, an enhanced version of GA by utilizing mutation and crossover for the solution of CHPED problems was adopted by [25]. They stated that primary GA is grown in three phases. In their first phase, they did not include the process of selection to bypass population diversity loss, while in the second phase, they utilized two various crossover operations to dig data about parents and produce possible children. In the third phase, they used the operation of mutation to substitute children with children of other parents. They proved that their Enhanced algorithm is the best substitute for the CHPED problem. Another study [26], was carried out to propose a novel quantum bat algorithm (QBA) based on quantum computing; its main purpose was to solve the problem of multi-objective combined economic emission dispatch (CEED). To minimize the system's nonlinearities, they represented CEED by utilizing the function of the cubic criterion. Their primary concern was the eruption of $CO_2$, NOx, and NOx and load dispatch. Thus, it is known as a multi-objective problem of optimization. Their outcomes proved that QBA is the best solution for the problem of CEED as compared to different available solutions. An improved self-adaptable differential evolution algorithm integrating with multiple mutation strategies (ADE-MMS) as a solution to ELD problems [27]. They suggested a strategy to improve and explore of basic DE problem. Their Enhanced method has 3 expansions of DE. Further, they suggested an approach to manage constraints of equality of problems of ED. Their algorithm enhances the speed of convergence as well as maintains a balance between exploration and exploitation. ADE-MMS is proved by them to be the most suitable solution.

Novel Differential Evolution algorithm was Enhanced in a study [28], for the solution of simultaneous power flow OARPD problems for renewable generators. Their Enhanced algorithm i.e., DEa-AR utilized a combination of arithmetic crossover and performed scaling based on Laplace distribution. For evaluation of their approach, they utilized the IEEE 57-bus system in various situations. The outcomes of their simulations verified that the suggested method could be used for solving OARPD problems with sources that are efficiently renewable and can provide optimum solutions. Qiao & Liu, [29] have carried out to propose a combined framework of EVs and wind farms (WEV) that reduced the over and underestimation of wind power by utilizing the discharging and charging capability of EVs. They designed a dynamic economic emission dispatching based on the WEV system (WE_DEED). They utilized their algorithm for the solution of complicated problems of WE_DEED. While they handle the limitations of WE_DEED through their Enhanced algorithm. They verified their algorithm on various 10 unit systems.

Authors enhanced a novel technique [30], for thermal plants' dispatch generating powers based on motion optimization algorithm (IMA). They achieved ELD being an objective function through implementing IMA. For the testing phase, multiple instances of various units of thermal plants were utilized to examine the execution of their algorithm. Their preceding outcomes were matched with various approaches.

A detailed description of related ELD applications concerning different evolutionary approaches is given in Table 1.

**Table 1** Detailed description of related ELD applications concerning different evolutionary approaches.

| Sr. | Ref | Proposed Technique | Dataset |
|---|---|---|---|
| 1 | [7] | BAT algorithm | - |
| 2 | [26] | Quantum bat algorithm (QBA) | - |
| 3 | [31] | Artificial bee colony algorithm | - |
| 4 | [32] | Bat Algorithm (BA) and Artificial Bee Colony (ABC) with Chaotic based Self-Adaptive (CSA) search strategy (CSA-BA-ABC) | 23 benchmark function and three CHPED problems |
| 5 | [25] | Improved genetic algorithm using novel crossover and mutation (IGA-NCM) | - |
| 6 | [33] | Learner Non-dominated Sorting Genetic Algorithm (NSGA-RL) | 10 famous multi-objective functions |
| 7 | [34] | Chaotic-crisscross differential evolution (CCDE) | Generalized test functions and two |

|  |  |  |  practical hydrothermal system problems |
|---|---|---|---|
| 8 | [35] | Differential Evolution algorithm (DEA) | IEEE-30 bus system |
| 9 | [29] | Dynamic economic emission dispatching based on WEV system (WE_DEED) | 10 unit systems. |
| 10 | [27] | Self-adaptable differential evolution algorithm integrating with multiple mutation strategies (ADE-MMS) | 4 DE algorithms are tested on the ten ELD problems with diverse complexities |
| 11 | [28] | Differential Evolution the algorithm denoted as DEa-AR | IEEE 57-bus system |
| 12 | [32] | Modified crow search algorithm (MCSA) | Five different well-known test systems |
| 13 | [36] | Multi-objective Multi-Verse Optimization algorithm | 140 bus system |
| 14 | [24] | Multi-objective economic and environmental dispatch problem (EEDP) | Five generation systems |
| 15 | [37] | Coyote Optimization Algorithm (COA) | Power system consisting thermal generator |
| 16 | [30] | Motion optimization algorithm (IMA) | Several cases of Different units of thermal plants |

## 3 Methodology

3.1 Problem Formulation

Emission can be included in economic dispatch's formulation in a variety of methods. Combined economic and emission dispatch (CEED) is one of the methods that are expressed as a problem of multi-objective optimization used to reduce emission and fuel costs to satisfy demand and avoid losses [38].

*3.1.1 Combined Emission and Economy Dispatch (CEED)*

CEED problem can be expressed as [39].

$$\sum_{i=1}^{N} P_i - P_l - P_d \tag{1}$$

$$FC = \sum_{i=1}^{N} a_i P_i^2 + b_i P_i + c_i \tag{2}$$

$$EC = \sum_{i=1}^{N} a_i P_i^2 + b_i P_i + c_i \,|Cost\ function \tag{3}$$

$$P_l = \sum_{i=1}^{N} \sum_{j=1}^{N} B_{ij} P_j P_i + \sum_{j=1}^{N} B0_i P_i + B_{00} \tag{4}$$

where $P_d$ represents total load demand, $P_l$ shows total transmission loss and $P_i$ is the power produced by $ith$ generator. Total fuel cost is denoted by FC and total emission is denoted with EC. In equation 2, $a_i$, $b_i$ and $c_i$ represent fuel cost coefficients. From eq 3, the cost function depends on the problem nature, it can be quadratic, square, sinusoidal, or any other function.

Referred to equation 4, $B_{ij}$ Coefficients or load flow can be utilized to find transmission losses denoted as $P_l$. Where $B0_i$ is coefficients vector of $B_{ij}$ and a value $B_{00}$. The price penalty factor is utilized to transform the problem of multi-objective optimization into a single-objective optimization problem as follows:

$$f(FC, EC) = Minimize\ (FC + EC) \tag{5}$$

Each plant or price penalty factor can be found for a specific demand as follows:

i. The ratio between the average fuel cost and the average emission of the maximum power capacity of that plant is found:

$$b_i = \frac{FC_i(U_i)}{EC_i(U_i)}, i = 1,2,n \tag{6}$$

where $(U_i)$ is ith unit of plant capacity.

ii. Plants are arranged in ascending order based on the value of the price penalty factor.

Cite as: Tahir, Barzan Hussein, Rashid, Tarik A., Rauf, Hafiz Tayyab, Bacanin, Nebojsa, Chhabra, Amit, Vimal, S., Yaseen, Zaher Mundher(2022). Improved Fitness-Dependent Optimizer for Solving Economic Load Dispatch Problem. Computational Intelligence and Neuroscience. Hindawi. DOI: https://doi.org/10.1155/2022/7055910    iii. Each unit ($U_i$) maximum capacity is added one at a time, beginning from the lowest value of $b_i$, unit until $\sum P_i \geq P_d$.

    iv. At this point bi, linked with the last unit of the process is the price penalty factor 'b', Rs/Kg for the provided demand of the load.

### 3.1.2 Emission Controlled Economic Dispatched (ECED)

Emission-controlled economic dispatch (ECED) is another method to reduce the economy related to a specific limit of emission concerning a specific demand. ECED problem's primary concern is to discover the cost-effective placement of plants while fulfilling the losses and demand and keeping the permissible limit of emission; FC needs to be reduced directed to equation 1, i.e., power balance constraint and emission limit constraint expressed as equation 7.

$$f(FC) = \sum_{i=1}^{N} P_i - P_l - P_d , \; P_l \leq P_i \leq P_d, EC \leq E_{limit} \qquad (7)$$

Here the system's total emission limit is denoted by $E_{limit}$.

### 3.2 Bee Swarming

This extraordinary insect is one of the most remarkable creatures since the old times. Honeybees have been the topic of scientific research. Moreover, multiple books and experiments have been carried out about honeybees; for instance, Ribbands published "Behavior and the Social Life of Honeybees" in 1953. "Anatomy of the Honey Bee " was written by Snodgrass in 1956, and Thomas D. Seeley wrote "the wisdom of hive" in 1995. The anatomy of a bee is shown in Figure 1. A process known as swarming is carried out to form new honeybee colonies:

    i. The old colony is left by queen bees among some workers and scout bees.

    ii. A swarming cycle is shown in Figure 2. A collection of thousands to tens

Page 11 of 42

      of thousands of bees make up aswarm.

  iii.    They make a cluster around the queen in some branch or a tree and twenty-fifty scouts are sent out todiscover some new proper hives.

  iv.    Finally, under the supervision of the scout's bees, all other bees fly to the new hive.

Scout bees check a hive for various standards to meet, such as, it must be wide enough to hold the entire swarm, the entrance must be small and must be at the bottom of the hive, and must get a particular amount of heat from sunlight [40]. Processes of decision-making of scout bees are the source of inspiration. When they find various proper hives, they select the best among them. The source of communication between scouts is the movement of their wings and legs, which is called bee dance. The new hive is selected after the agreement of 80% of the scouts [40].

In terms of algorithms, every hive utilized by a scout demonstrates a feasible solution of an artificial search agent, while the fittest hive expresses a global optimum solution which is represented in Table 2. The characteristics of the hive, like its size, size of entrance, location of the entrance can be viewed as a solution's fitness function. The process of collective decision-making of scouts is expressed as fitness weight *(fw)* in the algorithm.

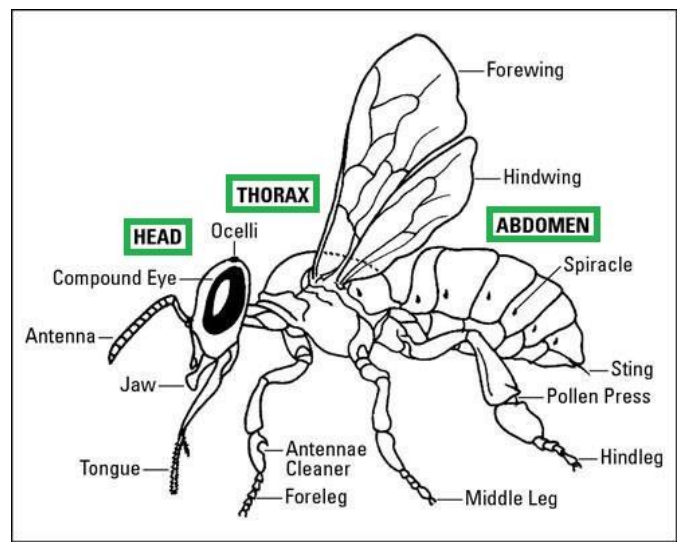

Fig. 1: Honey Bee Anatomy [17].

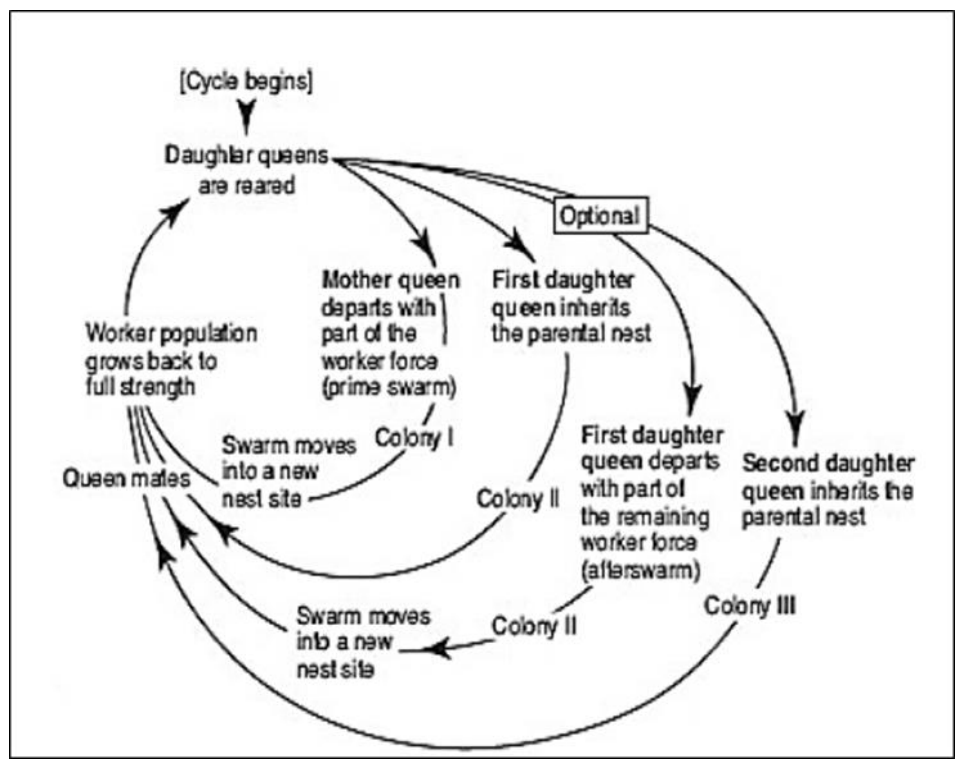

Fig. 2 Bee Swarming Process Cycle [17].

Table 2 FDO related Bee Biological characteristics.

| Sr. | Nature | Algorithm |
| --- | --- | --- |
| 1 | Selected hive | Global Solution |
| 2 | Scout collective decision | Objective Weight |
| 3 | Hive specification | Objective function |

| 4 | Hive | Solution found |
|---|---|---|
| 5 | Scout bee | Search agent |

### 3.3 Fitness Dependent Optimizer

The reproduction process of swarm bees is replaced by this algorithm. The major portion of the algorithm is obtained from hive exploring the process of scout bees from a pool of suitable options. The algorithm starts with the random initialization of the artificial scout population within the search space of $X_i (i = 1,2,...n)$; the position of every scout expresses a recently recognized hive. Scout bees keep on finding the more suitable hive; once they find a better hive, they neglect the previous better hive; the same is the case with the algorithm. Whenever it discovers a new, more suitable solution than the earlier determined solution is neglected. If they cannot find any other better solution than the previous one, they will consider the current solution as the best solution.

A mechanism of fitness weight and random walk is used to randomly explore the landscape by artificial scouts in this algorithm. The following equation expresses the movement of artificial scout bees.

$$X_{i,t+1} = X_{i,t} + pace \qquad (8)$$

Here $i$ denotes the current search agent, x denotes an artificial scout bee (search agent), $pace$ denotes the direction and movement rate, and $t$ denotes the current iteration of the artificial scout bee. $pace$ usually depends on $fw$ i.e. fitness weight while $pace$'s direction fully depends on a random mechanism. Therefore, minimization problems' $fw$ can be measured as:

$$fw = \left| \frac{x^*_{i,t\ fitness}}{x_{i,t\ fitness}} \right| - wf \qquad (9)$$

The current best global solution's fitness function value is denoted by $x^*_{i,t\ fitness}$, current solution's fitness function value is denoted by $x_{i,t\ fitness}$, the weight factor is expressed as $wf$ which can have only 0 or 1 value and is used to control $fw$. If $wf=1$ then it shows a low possibility of coverage and a high level of convergence. But if it is equal to 0, then it will not have any effect on equation (3.9), so it can be ignored, if the variable is $wf=0$ it will present us a more stable search. However, it reverses as the value of the fitness function entirely depends on the optimization problem. But, the value of $fw$ must be in the range of [0, 1]; still, in some situations when $fw = 1$, for instance, if the recent solution is the global best solution or global best solution and the recent solution is same or hold similar fitness value. Furthermore, a possibility exists when $fw = 0$, if $x^*_{i,t\ fitness} = 0$. Lastly, it must bypass the chances to divide a number with 0 in case $x_{i,t\ fitness}$, so, it must follow the following rules:

$$\begin{cases} fw = 1\ or\ fw = 0\ or\ x_{i,t\ fitness} = 0, pace = x_{i,t} * r & (10) \\ fw > 0\ and\ fw < 1 \begin{cases} r < 0, pace = (x_{i,t} - x^*_{i,t}) * fw * -1 & (11) \\ r \geq 0, pace = (x_{i,t} - x^*_{i,t}) * fw & (12) \end{cases} \end{cases}$$

where, a random number with a range of [-1, 1] is denoted by $r$. The random walk can be implemented in a variety of ways but here Levy flight is selected as its good distribution curve offers stable movements [22]. According to FDO mathematical complexity: its time complexity for every iteration is $O(p * n + p * CF)$, here $p$ denotes the size of the population, problem dimensions are denoted by $n$, and cost of the objective function is $CF$. Space complexity for every iteration is $O(p * CF + p * pace)$, here pace denotes the best previous paces stored. From this point, the time complexity of FDO is proportional to the number of iterations. Though, space complexity will remain identical throughout the sequence of iterations. For the calculation of objective value, FDO owns a simple tool for calculations; it only calculates one random number and fitness weight for every agent [41]. Similarly, DA alignment, attraction, separation, some random values, and distraction are required to be calculated, while a majority of them are accumulative and the value of one depends on the value of others making the calculations complicated [42].

### 3.3.1  Single-objective optimization-based FDO

FDO with single-objective optimization problems (FDOSOOP) starts with the initialization of artificial scouts on random locations of search landscape by utilizing lower and upper boundaries. For each iteration. It selects the global best solution, after that each artificial scout bee is computed by using equation (9). Then the value of $fw$ is examined to decide whether it is 1 or 0and if $x_{i,t\ fitness}$=0. $pace$ is generated by utilizing equation (10). But, a random number denoted by $r$ of range [-1, 1] will be generated if $fw>0\ and\ fw<1$. For calculation of $pace$ Equation (10) will be utilized if the value of $r$ is less than 0 and value of $fw$ will have a negative sign, however, for $r≥0$, the pace will be calculated with the help of equation (12) and $fw$ will have a positive sign. Random selection of signs for $fw$ will ensure the random search of artificial bees in all directions.

In FDO, direction and size of pace are controlled by the randomization method; however, only the direction of pace is usually controlled by this method; in such situations, the pace's size depends on $fw$. Whenever scout bees find a new solution, it is compared with the current solution to determine if it is better or not based on a fitness function. The earlier solution is neglected if, the better latest solution is obtained. Similarly, if it is not better than the previous value of pace, it will be used by the scout bee to continue. On the other hand, if a better solution cannot be achieved by utilizing the previous value of $pace$, then the current solution will be continued by FDO to the next iteration. In FDO, whenever a solution is acquired, the value is saved for utilization in the next iteration. Two minor alterations are required for the implementation of FDO in maximization problems. Equation (9) should be replaced by Equation (13) as it is the inverse variant of Equation (9).

$$fw = \left|\frac{x_{i,t\,fitness}}{x^*_{i,t\,fitness}}\right| - wf \tag{13}$$

Then the criteria for the selection of the best solution must be altered. The statement "if $(X_{t+1,i}fitness < X_{t,i}fitness)$" needs to be replenished with "if $(X_{t+1,i}fitness > X_{t,i}fitness)$".

### 3.3.2  Multi-objective optimization-based FDO

Multi-objective optimization problems (FDOMOOP) FDO begins with the initialization of artificial scouts into two-dimensional search space $(X_i, Y_i)$. Each scout bee in the search space of $(X_i, Y_i)$ can be defined as $X_i(i = 1,2,...n)$ and $Y_i(i = 1,2,...n)$. Then the value of $fw$ is examined to decide whether its 1 or 0 and if $x_{i,t\,fitness} = 0$ or $y_{i,t\,fitness} = 0$. In both cases, the $pace$ can be generated as $fw = \left|\frac{x_{i,t\,fitness}}{x^*_{i,t\,fitness}}\right| - wf$ and $fw = \left|\frac{y_{i,t\,fitness}}{y^*_{i,t\,fitness}}\right| - wf$.

### 3.4 Enhanced Method

Multi-dimensional and multi-objective optimization algorithms tend to perform better for solving the linear and non-linear constraint problems than the single-dimensional and single-objective optimization algorithms, especially in the case of ELD, when both the emission rate and fuel cost need to be minimized to approach the total loss and power demand. FDO is a multi-objective metaheuristic algorithm and, therefore, best suitable for solving constrained ELD problems. It can be implied to complex problems with non-linear approximation. In this thesis, the authors have carried out FDO to solve the ELD problem by minimizing fuel cost, emission allocation, and Transmission Loss. Besides, authors have employed a novel variant of FDO which incorporates novel population initialization techniques and employed sine maps to select the weight factor for FDO dynamically.

### *3.4.1 Population Initialization*

The swarm or group of swarms needs to be fired randomly to obtain their initial fitness solution in the optimization process. The entire process is called population initialization. The most conventional method to assign an initial location to each individual is through a random number generator following the normal distribution. However, the major drawback to using the random number generator is premature convergence and abnormal exploration and exploitation. The random number generator developed a random number between the internal of 0 and 1. The probability of obtaining an optimal solution in the case of local minima is reduced when the initial locations are directed far from the solution, and each individual requires more steps and iterations to seek the entire solution. The swarm can be stuck into premature convergence during the searching process and lead to poor exploration. Similarly, as opposed to this, the probability of obtaining a global solution in the case of global minima is diminished when the primary positions are delivered too near around the search space while the solution is out of search space; hence each individual requires more steps and iterations to seek the entire solution and can be stuck into the premature convergence and leads to poor exploitation.

### *3.4.2 Quasi-Random Sequence Initialization*

Quasi-Random is a distinction of n-rows that occupies n-dimensional search space. It is also called a low-disparity sequence. However, the usual standard quasi-random sequences and odd numbers all give consistently suitable sequences. There is a significant distinction between these two patterns, aside from the standardized manner. An identical arbitrary generator on (0, 1) will deliver sequences, so every preliminary has a similar likelihood of producing a point on equivalent subintervals, for instance [(0, 1/2), (1, 1/2)]. In this manner, it is attainable for n preliminaries to inadvertently all extend in the top half of the range, while the (n+1) points fall inside the other of the two parts with a likelihood of 1/2. While this is not the situation with the quasi-random sequences, the generated sequences are obliged by a low-inconsistency prerequisite that has a net impact on centers being created in a profoundly connected way. To avoid the premature convergence problem in FDO, authors have carried out one of the quasi-random sequences called the Sobol sequence for the population's initialization.

### 3.4.3 Quasi-Random Sequence Initialization

Sobol sequence is a low discrepancy sequence that was first Enhanced by mathematicians in Russia in 1967 [43]. It mimics the random distribution by appropriating a base of two to shape progressively better uniform edges of the required interval and afterward reorder the directions in each measurement. Following are prime steps to generate the Sobol sequences $S^d$

i. Let $S^d$ be the hypercube with the interval of $S^d = [0,1]^d$ and $d$-dimensional. The approximation function $f^{opr}$ is integrated over the hypercube $S^d$.

ii. The Sobol sequence termed as $Sobol\ [x, y]$ can be generated using the following equation over the nonlinear approximation of $S^d$.

$$\lim_{x \to \infty} \frac{1}{x} \sum_{i=1}^{x} f(S_i) = \int_{S^d}^{i} f \qquad (14)$$

iii. It is a notable pattern against each dimensional vector that for the whole to reach towards the indispensable points $S^d$. Furthermore, the second great feature would be that the forecasts of $x$ in the low range of the dimensioned face of $S^d$ cover most of the search area in terms of optimization.

iv. Subsequently, the comparable center of $S^d$ does not meet the criteria because in lower measurements numerous focuses will be at a similar spot, in this way unnecessary for the vital estimation.

The comparison of FDO population initialization with random numbers following the Sobol distribution and the uniform distribution is presented in Figures 3 and 4 respectively. The following equation is used in the standard FDO for the swarm to select their initial locations to seek the entire optimal solution.

$$Random_i(i = 1,2,...n) \therefore [0,1] \qquad (15)$$

In the Enhanced FDO, authors have selected the interval of [0, 1] for generating both sequences uniform and Sobol sequences in the process of FDO population initialization. The Enhanced equation for initializing the swarm in FDO is presented below:

$$Sobol_i(i = 1,2,...n) \therefore [0,1] \qquad (16)$$

where [0,1] in Equations 15 and 16 represent the standard limits of both generated sequences. The uniform random positions can be seen in Figure 3 with very random locations and ill-patterned sequences which may lead to poor exploitation. As compared to the uniform random, the Sobol sequence comes up with a well-patterned sequence in Figure 4 which may lead the swarm to converge maturely.

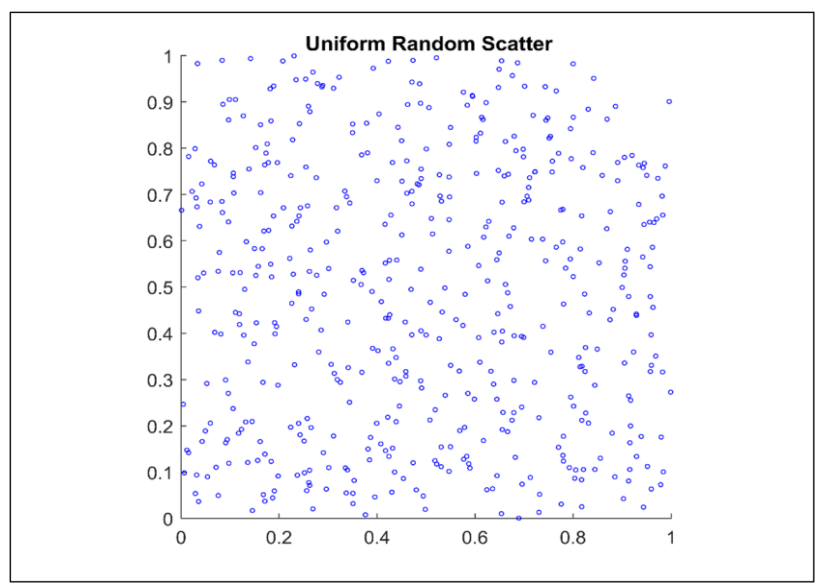

**Fig. 3** Population initialization with random number generator following the uniform sequence.

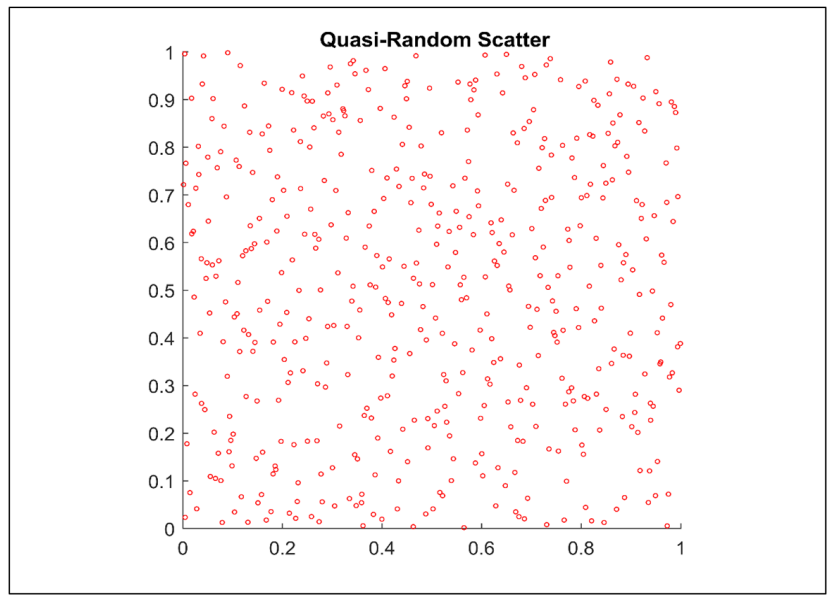

**Fig. 4** Population initialization with Sobol sequence following the random distribution.

### 3.4.4 Enhanced approach for updating Weight factor

The weight factor is revealed as $wf$ which can have particularly 0 or 1 utility and is utilized to control $wf$. If $wf$ belongs to absolute 1 then it confers a low probability of coverage and a high level of convergence. However, if the weight factor $wf$ belongs to 0, then it will not have any influence on equation 17, so it can be ignored, if the variable is $wf = 0$ it will present us a more stable search. However, it reverses as the value of the fitness function entirely depends on the optimization problem.

$$fw = \left|\frac{x^*_{i,t\ fitness}}{x_{i,t\ fitness}}\right| - 0 \tag{17}$$

However, in unusual circumstances, when the weight factor belongs to absolute one as shown in equation 18, for instance, if the new solution is the global best solution or local best solution and the new solution is identical or operates similar fitness value.

$$fw = \left|\frac{x^*_{i,t\,fitness}}{x_{i,t\,fitness}}\right| - 1 \qquad (18)$$

$wf$ should be balanced enough to control the exploitation and exploration for the controlled convergence rate when leads to absolute $0\,wf$ and absolute $1\,wf$. For this, the authors produce a chaotic effect by using sine maps to the weight factor $wf$ between the interval of [0,1].

### 3.4.5 Chaotic Sine Map

Chaotic maps produce uncontrolled groupings during the metaheuristic algorithm. The authors practiced the benefit of a chaotic sine pattern to update the weight factor [22]. The sine map is chaotic and used to produce a quarter effect between the interval of 0 and 1. When the weight factor becomes skewed towards 0, the sine wave covers the low balance and controls the low convergence rate. Similarly, when the weight factor becomes skewed towards 1, the sine wave covers the high balance and controls the high convergence rate. This phenomenon iteratively maintains the balance with optimal weight factor throughout the last epoch. A chaotic sign map can be defined as:

$$S_{map} = \frac{m}{4}\sin(\pi x_i) \qquad (19)$$

where $0 < m < 4$ is the controlling factor. The author chooses $m = 0.3$ with the most optimal sequence. In terms of weight factor the equation becomes:

$$w_s = \frac{m}{4}\sin(\pi wf) \qquad (20)$$

The Enhanced variant of FDO utilized the following equation to update the fitness weight $fw$.

$$fw = \left|\frac{x^*_{i,t\,fitness}}{x_{i,t\,fitness}}\right| - w_s \qquad (21)$$

The flowchart of the Enhanced FDO along the ELD application is presented in Figure 5.

# 4 Application Results

4.1 Data Set Overview

The performance of the Enhanced variant of FDO is evaluated through 24 units taken from the 18-unit system and 20 unit system with each of the 6 unit case study chunks by optimizing the fitness function enlisted in Equation 1. The parameter sets used in the experiment for each unit are listed below:

Total number of units used in the experiment = 24, Total power demand = 400, 700, Number of iterations = 100, 200, Population size = Number of Bee scouts = 50, the Beta coefficient used for 24 units according to each chunk of 6 units in the exploring capacity with a power demand of 400 MW and 700 MW are presented as follows:

$$Beta\ coefficient = 1e^{-4} \times \begin{bmatrix} 1.4 & .17 & .15 & .19 & .26 & .22 \\ .17 & .60 & .13 & .16 & .15 & .20 \\ .15 & .13 & .65 & .17 & .24 & .19 \\ .19 & .16 & .17 & .71 & .30 & .25 \\ .26 & .15 & .24 & .30 & .69 & .32 \\ .22 & .20 & .19 & .25 & .32 & .85 \end{bmatrix} \quad (22)$$

Detail of 24 units used to minimize the fuel cost, emission allocation, and Transmission Loss with 400 and 700 power load is presented in Table 2. In table 2, $Pmin$ and $Pmax$ represent the lower and upper plant limits. Where other parameters can be defined as $a = \frac{\$}{MW^2}, b = \frac{\$}{MW}, c = \$$.

The data set used for the simulation consists of two chunks with 12 generating thermal units each. The first 12 generating units (1 to 12) are taken from Sys_18 U with all plant limitations and beta coefficients as represented in Equation (22).

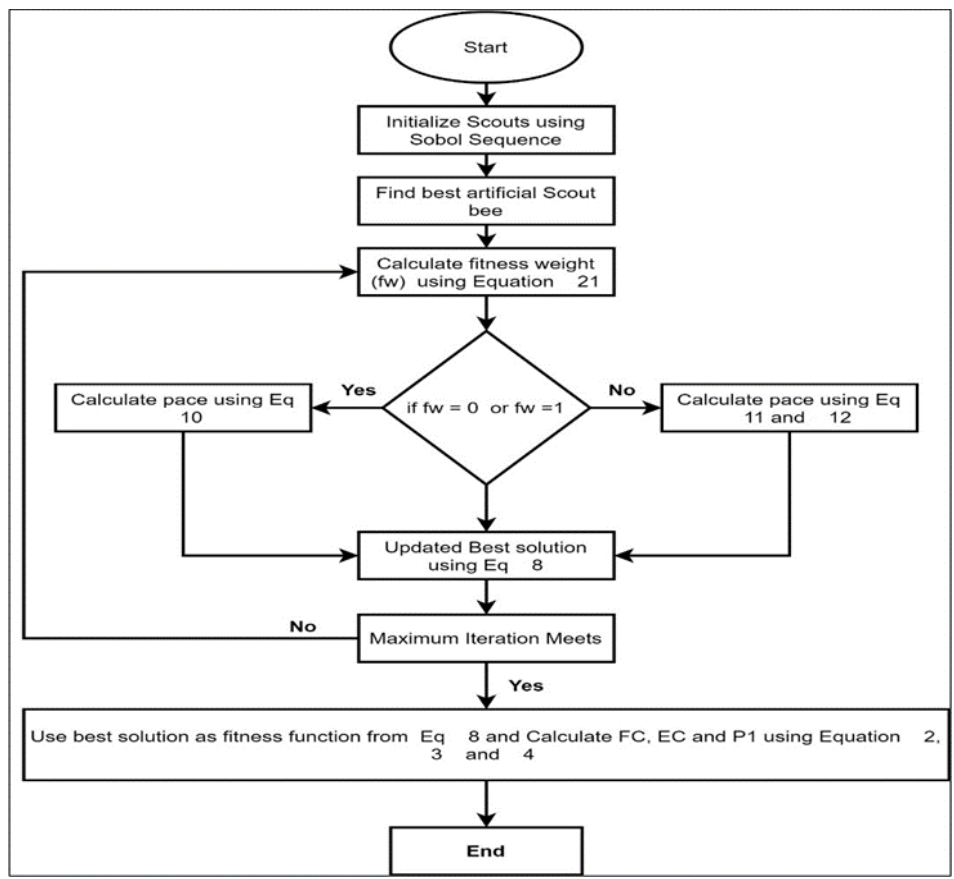

**Fig. 5** Flowchart for Enhanced FDO algorithm along with ELD application.

Similarly, the last 12 generating thermal units (13 to 24) are taken from Sys_20 U with all plant limitations and beta coefficients as described in Equation (22). All thermal units employed for the empirical analysis are ramp-limits-free and do not endure in the prohibited zone for the smooth objective function. Comparison of simulation results on the ELD problem (FDO vs Enhanced FDO) with nonlinear optimization on 100 epochs with a power demand of 400 and 700 are presented in Tables 3 and 4. The Total fuel error and the transmission cost are the minimum global fitness achieved by optimizing the ELD problem as Minimize $f(FC, EC), \ni, \sum_{i=1}^{N} P_i = P_d + P_l, L_i \leq P_i \leq U_i$. Similarly, in Table 4 and 5, a Comparison of simulation results on the ELD problem (FDO vs Enhanced FDO) with nonlinear optimization on 200 epochs with a power demand of 400 and 700 are presented. Certainly, for our problem statement, the accentuation is to distinguish which epoch setting requires the most minimal fuel cost and Transmission Loss to discover arrangements of a specific worthy quality. Furthermore, the Power demand is also analyzed in a roundabout way, to give in any event a complex reflection of the complexities of the various calculations considered in our relative examination.

The obtained results certainly take the fact that the population initialization and optimal fitness factor of FDO make some impact on the global best of FDO as compared to the Enhanced variant of FDO in terms of optimal allocation emission. The reason behind developing each thermal generation chunk with 6 units is to investigate the impact of greater dispersion on the total fuel cost and minimum error. To visualize the obtained results emission allocation results, a Convergence comparison of FDO with the Enhanced variant of FDO on the first 6 thermal units with 100 epochs and different power demands are illustrated in Figure 6.

**Table 3** Twenty-four units used with a chunk of 6 units in the exploring capacity with a power demand of 400 MW and 700 MW.

| Units | $P_{min}$ | $P_{max}$ | a | B | C |
|---|---|---|---|---|---|
| 1 | 7 | 15 | 0.602842 | 22.45526 | 85.74158 |
| 2 | 7 | 45 | 0.602842 | 22.45526 | 85.74158 |
| 3 | 13 | 25 | 0.214263 | 22.52789 | 108.9837 |
| 4 | 16 | 25 | 0.077837 | 26.75263 | 49.06263 |
| 5 | 16 | 25 | 0.077837 | 26.75263 | 49.06263 |
| 6 | 3 | 14.75 | 0.734763 | 80.39345 | 677.73 |
| 7 | 3 | 14.75 | 0.734763 | 80.39345 | 677.73 |
| 8 | 3 | 12.28 | 0.514474 | 13.19474 | 44.39 |
| 9 | 3 | 12.28 | 0.514474 | 13.19474 | 44.39 |
| 10 | 3 | 12.28 | 0.514474 | 13.19474 | 44.39 |
| 11 | 3 | 12.2 | 0.514474 | 13.19474 | 44.39 |

|    |     | 8   |          |          |           |
|----|-----|-----|----------|----------|-----------|
| 12 | 3   | 24  | 0.657079 | 56.70947 | 574.9603  |
| 13 | 150 | 600 | 0.00068  | 18.19    | 1000      |
| 14 | 50  | 200 | 0.00071  | 19.26    | 970       |
| 15 | 50  | 200 | 0.0065   | 19.8     | 600       |
| 16 | 50  | 200 | 0.005    | 19.1     | 700       |
| 17 | 50  | 160 | 0.00738  | 18.1     | 420       |
| 18 | 20  | 100 | 0.00612  | 19.26    | 360       |
| 19 | 25  | 125 | 0.0079   | 17.14    | 490       |
| 20 | 50  | 150 | 0.00813  | 18.92    | 660       |
| 21 | 50  | 200 | 0.00522  | 18.27    | 765       |
| 22 | 30  | 150 | 0.00573  | 18.92    | 770       |
| 23 | 100 | 300 | 0.0048   | 16.69    | 800       |
| 24 | 150 | 500 | 0.0031   | 16.76    | 970       |

Moreover, Convergence comparison (Transmission Loss) of FDO with the Enhanced variant of FDO on the 24 thermal units with (100, 200 epochs) and different power demands are demonstrated in Figures 7 and 8. To validate the achieved results on the ELD problem (FDO vs Enhanced FDO), this research used an ANOVA test. The main reason behind performing the ANOVA test is to find the significant difference between the standard and Enhanced FDO in terms of minimization. The additional reason to perform the ANOVA test is to determine which parameter delivers outcomes with critical contrasts, considering the target value accomplished by Enhanced FDO from each run of the considerable number of tests performed. Graphical representation of one-way ANOVA Test comparison (optimal allocation emission) of FDO with the Enhanced variant of FDO on the 24 thermal units with (100,200 epochs) and several power demands are illustrated in Figures 8 and 9, respectively.

**Table 4** Comparison of simulation results on the ELD problem (FDO vs Enhanced FDO) with nonlinear optimization on 100 epochs and 400 power demand. The optimal values are exhibited in boldface.

| **Units** | **Power Demand = 400** |
|-----------|------------------------|

|  | **Optimal Allocation Emission (*lbs*)** | |
|---|---|---|
|  | **FDO** | **Enhanced FDO** |
| 1 | 70.44063664 | **69.91458228** |
| 2 | 69.28036315 | **68.72201216** |
| 3 | 38.43849912 | **37.87285707** |
| 4 | 31.18554733 | **30.67657178** |
| 5 | 31.07224457 | **30.56379958** |
| 6 | 160.2587123 | 162.2504561 |
| **Total Fuel Cost ($)** | 2.05E+05 | **2.04E+05** |
| **Transmission Loss** | 0.676 | **2.79E-04** |
| 7 | 111.6951678 | **111.360789** |
| 8 | 44.39 | 44.39 |
| 9 | 44.39 | 44.39 |
| 10 | 44.39 | 44.39 |
| 11 | 44.39 | 44.39 |
| 12 | 111.431507 | **111.0794924** |
| **Total Fuel Cost ($)** | 1.05E+05 | **1.04E+05** |
| **Transmission Loss** | 0.6867 | **2.81E-04** |
| 13 | 18.32981341 | **18.24689868** |
| 14 | 59.0172677 | **58.85809285** |
| 15 | 58.93086524 | **58.77200905** |
| 16 | 58.7043227 | **58.5458324** |
| 17 | 58.92968203 | **58.77222232** |
| 18 | 146.7176828 | 146.8052033 |
| **Total Fuel Cost** | 1.25E+06 | 1.24E+06 |
| **Transmission Loss** | 0.6296 | **2.59E-04** |
| 19 | 122.1815342 | **122.1372312** |
| 20 | 62.6886485 | **62.53294149** |
| 21 | 62.13776388 | **61.98118821** |
| 22 | 103.4629522 | **103.3564131** |
| 23 | 30.48017867 | **30.36716223** |
| 24 | 19.70886257 | **19.62533477** |
| Total Fuel Cost ($) | 1.31E+06 | **1.30E+06** |
| Transmission Loss | 0.6599 | **2.71E-04** |

**Table 5** Comparison of simulation results on the ELD problem (FDO vs Enhanced FDO) with nonlinear optimization on 100 epochs and 700 power demand. The optimal values are exhibited in boldface.

| Units | Power Demand = 700 | |
|---|---|---|
| | Optimal Allocation Emission (*lbs*) | |
| | **FDO** | **Enhanced FDO** |
| 1 | 85.7416 | **85.74158** |
| 2 | 85.7416 | **85.74158** |
| 3 | 108.9837 | **83.36202838** |
| 4 | 49.0626 | 49.06263 |
| 5 | 49.0626 | 49.06263 |
| 6 | 525.6109 | **347.0304788** |
| Total Fuel Cost ($) | 6.51E+05 | **6.46E+05** |
| Transmission Loss | 2.2609 | **9.27E-04** |
| 7 | 259.65771 | **258.266736** |
| 8 | 44.39 | 44.39 |
| 9 | 44.39 | 44.39 |
| 10 | 44.39 | 44.39 |
| 11 | 44.39 | 44.39 |
| 12 | 265.4517495 | **264.1743532** |
| Total Fuel Cost ($) | 4.50E+05 | **4.45E+05** |
| Transmission Loss | 2.6695 | **0.0011** |
| 13 | 33.35092403 | **33.16801208** |
| 14 | 104.2882238 | **103.890753** |
| 15 | 104.0208305 | **103.6249588** |
| 16 | 103.3245773 | **102.9315033** |
| 17 | 103.1925346 | **102.8020805** |
| 18 | 253.7447758 | **253.5834798** |
| Total Fuel Cost ($) | 3.73E+06 | **3.71E+06** |
| Transmission Loss | 1.9219 | **7.88E-04** |
| 19  | 210.6754314 | **210.3359493** |
| 20  | 110.5273588 | **110.1212242** |
| 21  | 109.860002  | **109.4539346** |
| 22  | 181.2722755 | **180.8655902** |
| 23  | 54.11026706 | **53.84868486** |
| 24  | 35.56430084 | **35.37544006** |
| **Total Fuel Cost ($)** | 3.92E+06 | **3.90E+06** |
| **Transmission Loss** | 2.0096 | **8.23E-04** |

**Table 6** Comparison of simulation results on the ELD problem (FDO vs Enhanced FDO) with nonlinear optimization on 200 epochs and 400 power demand. The optimal values are exhibited in boldface.

| Units | Power Demand = 400 | |
| --- | --- | --- |
|     | **Optimal Allocation Emission** (*lbs*) | |
|     | FDO | Enhanced FDO |
| 1   | 69.91436113 | **69.91435917** |
| 2   | 68.72177747 | **68.72177539** |
| 3   | 37.87262247 | **37.87262039** |
| 4   | 30.67636117 | **30.67635931** |
| 5   | 30.56358919 | **30.56358733** |
| 6   | 162.251291 | 162.2512984 |
| **Total Fuel Cost ($)** | 2.04E+05 | 2.04E+05 |
| **Transmission Loss** | 2.45E-06 | **2.69E-12** |
| 7   | 111.3606533 | **111.360652** |
| 8   | 44.39 | 44.39 |
| 9   | 44.39 | 44.39 |
| 10  | 44.39 | 44.39 |
| 11  | 44.39 | 44.39 |
| 12  | 111.0793492 | **111.079348** |
| **Total Fuel Cost ($)** | 1.04E+05 | 1.04E+05 |
| **Transmission Loss** | 2.47E-06 | **2.72E-12** |
| 13 | 18.24686486 | **18.24686456** |
| 14 | 58.85802787 | **58.85802729** |
| 15 | 58.7719442 | **58.77194362** |
| 16 | 58.5457677 | **58.54576712** |
| 17 | 58.77215804 | **58.77215747** |
| 18 | 146.8052396 | 146.8052399 |
| **Total Fuel Cost ($)** | 1.24E+06 | 1.24E+06 |
| **Transmission Loss** | 2.27E-06 | **2.50E-12** |
| 19 | 122.1372134 | **122.1372132** |
| 20 | 62.53287799 | **62.53287743** |
| 21 | 61.98112435 | **61.98112378** |
| 22 | 103.3563698 | **103.3563694** |
| 23 | 30.36711614 | **30.36711573** |
| 24 | 19.62530071 | **19.62530041** |
| **Total Fuel Cost ($)** | 1.30E+06 | 1.30E+06 |
| **Transmission Loss** | 2.38E-06 | **2.62E-12** |

Table 7 Comparison of simulation results on the ELD problem (FDO vs Enhanced FDO) with nonlinear optimization on 200 epochs and 400 power demand. The optimal values are exhibited in boldface.

| Units | Power Demand = 700 | |
|---|---|---|
| | Optimal Allocation Emission (*lbs*) | |
| | **FDO** | **Enhanced FDO** |
| 1 | 85.74158 | 85.74158 |
| 2 | 85.74158 | 85.74158 |
| 3 | 83.36112205 | **83.36111401** |
| 4 | 49.06263 | 49.06263 |
| 5 | 49.06263 | 49.06263 |
| 6 | 347.0304661 | **347.030466** |
| **Total Fuel Cost ($)** | 6.46E+05 | 6.46E+05 |
| **Transmission Loss** | 8.15E-06 | **8.95E-12** |
| 7 | 258.266172 | **258.266167** |
| 8 | 44.39 | 44.39 |
| 9 | 44.39 | 44.39 |
| 10 | 44.39 | 44.39 |
| 11 | 44.39 | 44.39 |
| 12 | 264.1738376 | **264.173833** |
| **Total Fuel Cost ($)** | 4.45E+05 | **4.45E+05** |
| **Transmission Loss** | 9.57E-06 | **1.05E-11** |
| 13 | 33.16793762 | **33.16793696** |
| 14 | 103.8905912 | **103.8905898** |
| 15 | 103.6247976 | **103.6247962** |
| 16 | 102.9313433 | **102.9313419** |
| 17 | 102.8019216 | **102.8019202** |
| 18 | 253.5834156 | **253.583415** |
| **Total Fuel Cost ($)** | 3.71E+06 | 3.71E+06 |
| **Transmission Loss** | 6.92E-06 | **7.60E-12** |
| 19 | 210.3358119 | **210.3358107** |
| 20 | 110.121059 | **110.1210576** |
| 21 | 109.4537694 | **109.453768** |
| 22 | 180.8654252 | **180.8654237** |
| 23 | 53.84857844 | **53.8485775** |
| 24 | 35.37536322 | **35.37536254** |
| **Total Fuel Cost ($)** | 3.90E+06 | 3.90E+06 |
| **Transmission Loss** | 7.23E-06 | **7.94E-12** |

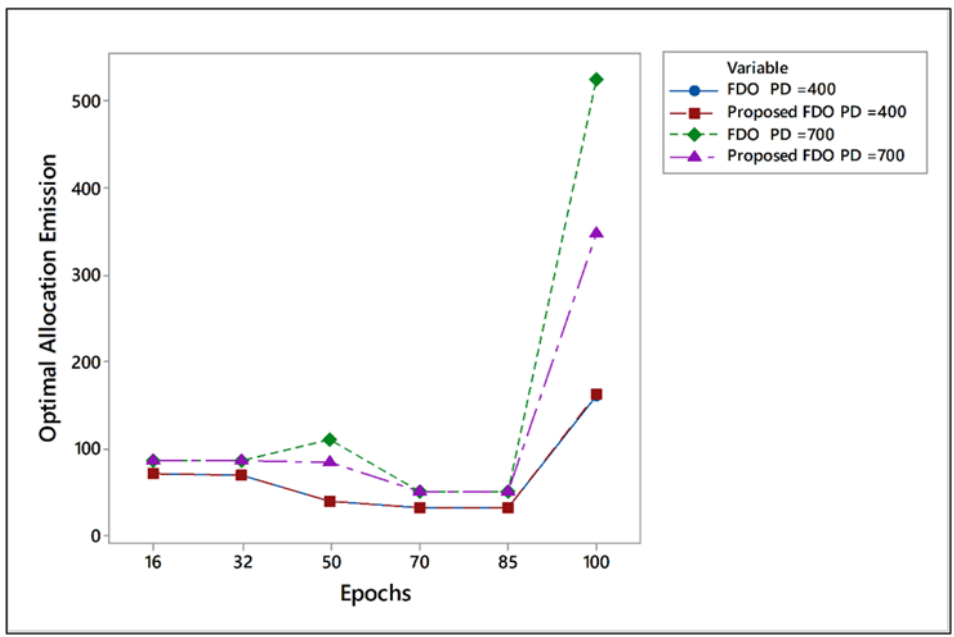

**Fig. 6** Convergence comparison (optimal allocation emission) of FDO with the Enhanced variant of FDO on the first 6 thermal units with 100 epochs and different power demands.

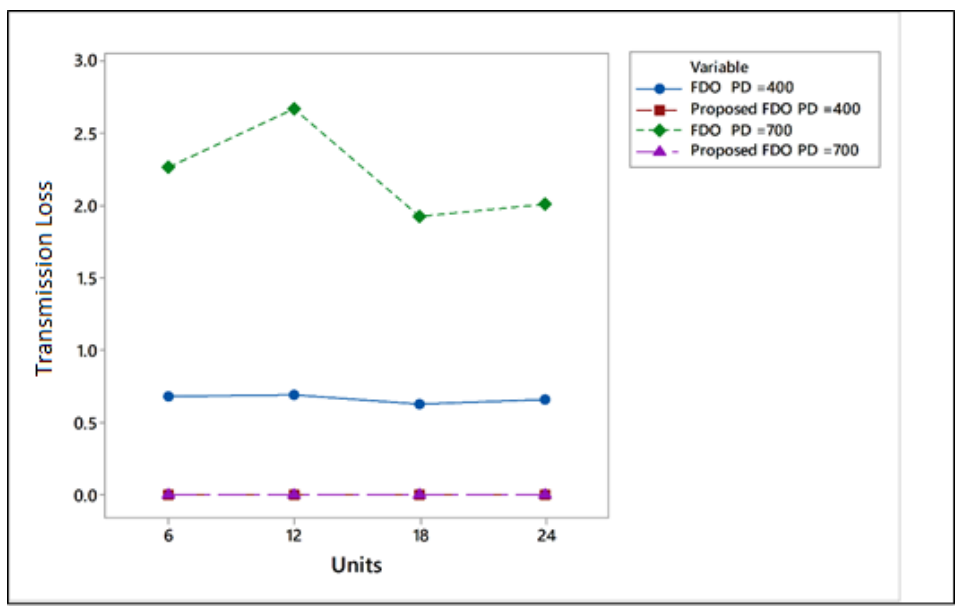

**Fig. 7** Convergence comparison (Transmission Loss) of FDO with the Enhanced variant of FDO on the 24 thermal units with 100 epochs and different power demands.

4.2 Optimal Allocation Emission

It is earlier mentioned that each chunk of thermal units are tested on 100 independent runs with 100 and 200 epochs considering two different combinations of power demand. The authors observed a significant improvement in the optimal power allocation generated by the Enhanced FDO for the first 6 thermal units as compared to the conventional FDO (Referred to Table 4). All 5 units' results obtained by Enhanced FDO outperformed FDO except the 6th unit with 162.2504561 optimal emission allocation on 100 epochs and 400 power demand. As contrasted to the first chunk of thermal units, the performance of the Enhanced algorithm was observed less when optimizing emission allocation. It can be seen from Table 4 that only thermal units 7 and 12 gained better emission rates, which lead to greater divergence of the whole population. However, thermal units 8, 9, 10, and 11 show equal empirical performance for both FDO and Enhanced FDO with a 44.3 emission rate.

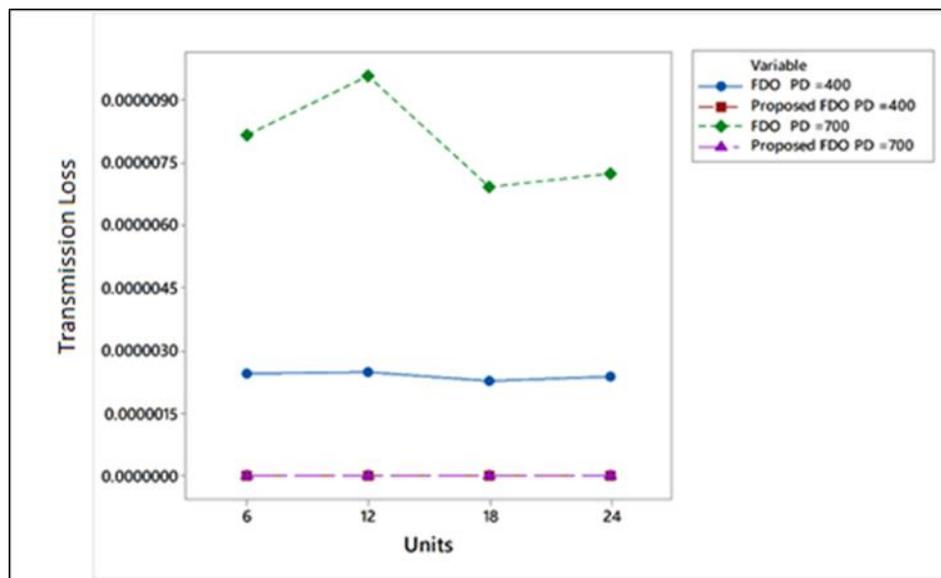

**Fig. 8** Convergence comparison (Transmission Loss) of FDO with the Enhanced variant of FDO on the 24 thermal units with 200 epochs and different power demands.

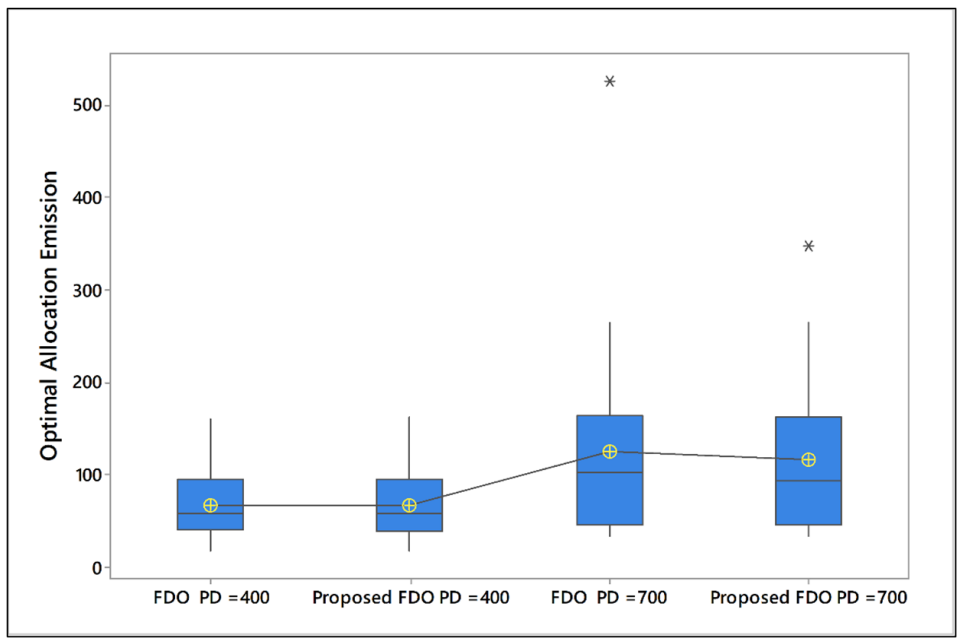

**Fig. 9** One-way ANOVA Test comparison (optimal allocation emission) of FDO with the Enhanced variant of FDO on the 24 thermal units with 100 epochs and different power demands.

In the case of the third chunk, thermal units 13 to 18, the emission allocation rate is not significantly improved using Enhanced FDO instead of the standard FDO. The Enhanced version obtained 18 with unit 13, 58 with units 14, 15, 16, 17, and 58 with unit 18, respectively, on 100 epochs and 400 power demand, which shows a slight improvement. This slight impact of Enhanced FDO reveals the impact of robust population initialization on the ELD emission allocation. Lastly, the fourth chunk of the thermal unit from Table 4 exhibits outstanding results of the Enhanced algorithm on the entire parameter setting except for the last thermal unit with a 19.62533477 emission rate.

From Table 5, when power demand raised 400 to 700, the Enhanced algorithm also improves the swarm convergence, and hence the optimal fitness factor works here. This phenomenon shows the inverse divergence of the global best computed with the Enhanced fitness factor, which leads to the emission allocation of the thermal units 3 and 6 from 68.72201216, 162.2504561 to 83.36202838 and 347.0304788, respectively. Similarly,

premature convergence is highly tackled by the Enhanced algorithm when seeing a significant decrease in the average optimal allocation for the first six thermal units. Tables 5 and 6 show similar convergence behavior of Enhanced FDO compared to the FDO using 200 epochs with 400 and 700 power demand. However, the optimal fitness factor produces less impact than the effect produced when testing on 100 epochs. This can be due to the dimension reduction that occurs in higher generations.

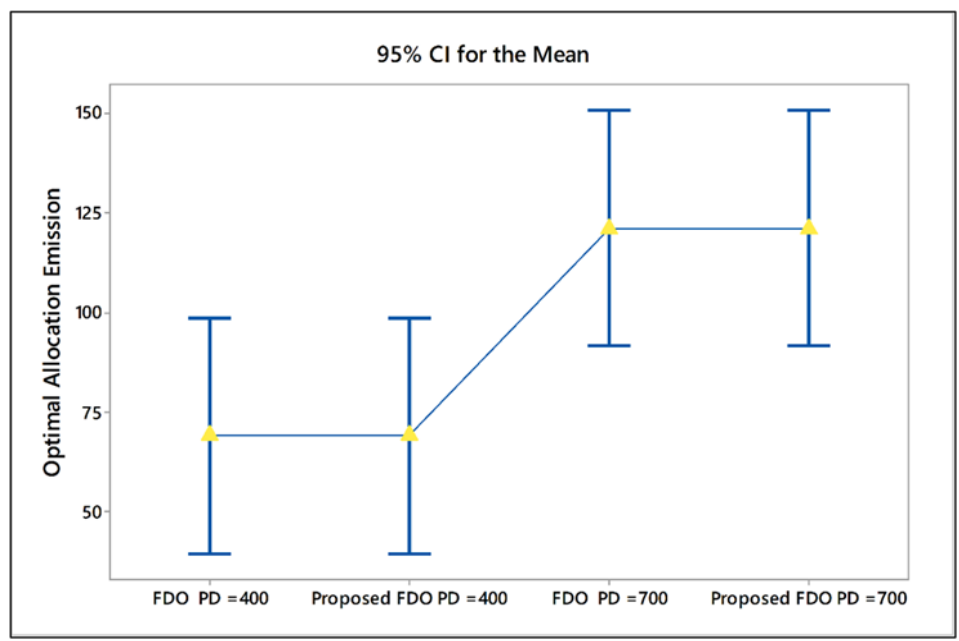

**Fig. 10** One-way ANOVA Test comparison (optimal allocation emission) of FDO with the Enhanced variant of FDO on the 24 thermal units with 200 epochs and different power demands.

4.3 Fuel Cost

Fuel cost minimization on the same hydro energy and power demand is a big issue when several units are mimicking in parallel. This can be optimized by considering the current fuel cost as the global best for each of the individuals in the FDO. However, the minimal risk is premature convergence which leads to double computing cases and wastage of time with the cost approximately equal to the standard fuel cost. Robust population initialization decreases the chance of premature convergence. Hence the Enhanced FDO used optimal fitness factors in combination with the Sobol operator to minimize the fuel cost.

62

63  It can also be observed in Tables 3 to 5 that the fuel cost difference between FDO and
64  Enhanced FDO is notable for all 24 thermal units on 100 and 200 epochs with a power demand
65  of 400 and 700 sequentially. However, this significant minimal difference in cost can impact the
66  whole unit generation cost. The trends for each chunk of the thermal unit from Table 4 to 6, confirm
67  the directly proportional relationship between the power demand and the fuel cost directly. The
68  minimization range is constant between them which explicates the strong divergence and influential
69  fitness factor. Increasing power demand will lead the fuel cost to increase with a constant
70  proportion of difference produced by FDO and Enhanced FDO.

71  4.4 Transmission Loss

72  Transmission error is essential for significant distance potential transmission and it grows with
73  an expansion in the measure of capacity to be dispatched. Therefore, the utilization of
74  inexhaustible force from the sustainable plants close to the heap focuses diminishes the
75  transmission losses. Appropriating the sustainable power sources all through the working time
76  frames as opposed to utilizing them during their accessible period will assist with diminishing
77  both expense and the transmission loss.

78      Compared to the optimal emission allocation and fuel cost, the authors have received
79  the best optimal transmission loss results. The Enhanced algorithm FDO reduced the loss with
80  a 60 % rate on average. The authors can perceive that in Table 4, FDO minimizes the loss to
81  0.676, 0.6867, 0.6296, and 0.6296 for four chunks of the thermal unit with 100 epochs and 400
82  power demand as compared to the Enhanced FDO, which reduces it to 2.79E-04, 2.81E-04,
83  2.59E-04, and 2.71E-04 for four chunks of the thermal unit with 100 epochs and 400 power
84  demand sequentially. Enhanced DFO significantly outperformed standard FDO for
85  minimization transmission loss.

86  Likewise, the Enhanced algorithm FDO decreased the loss by a 300 % rate regularly.
87  The study can comprehend that in Table 5, FDO minimizes the loss to 2.2609, 2.6695, 1.9219,
88  and 2.0096 for four chunks of the thermal unit with 100 epochs and 700 power demand as
89  contrasted to the intended FDO, which overcome it to 9.27E-04, 0.0011, 7.88E-04, and 8.23E-
90  04 for four chunks of the thermal unit with 100 epochs and 700 power demand sequentially.
91  Tables 6 and 7 explicitly formulate the same trends between standard and Enhanced FDO with
92  200 significant differences in Transmission error on 200 epochs and 400 and 700 power
93  demand. Figure 6 confirms the optimal convergence comparison in the case of optimal
94  allocation emission (FDO with the Enhanced variant of FDO0 on the first six thermal units
95  with 100 epochs and different power demands. Figures 7 and 8 dispense the clear-cut
96  Transmission Loss difference. Units are presented on the X-axis while Transmission Loss is
97  enlisted on Y-axis.

98  To validate the obtained results, ANOVA statistical analysis for Figures 9 and 10
99  reinforces the best performance of the Enhanced FDO algorithm and encourages the solution
100 for other constraints as well. The box representation for Enhanced FDO with 400 power
101 demand is proved as an optimal solution with optimal chunk. Similarly, the interval plot
102 representation for Enhanced FDO with 700 power demand is determined as an optimal solution
103 with an optimal chunk.

104 **5. Conclusion**

105 This research work introduced a variant of the FDO algorithm motivated by scout bees in the
106 hive exploring the process of seeking food from a pool of suitable options. The Enhanced
107 variant is utilized to solve the economic load dispatch problem. FDO and its modified version
108 are motivated to upgrade the minimization capability during weight optimization of economic
109 loads dispatch. Each individual of the scout bee is represented as output power generated
110 through each thermal unit. The study deals with three types of constraints in this work: power balance capacity, transmission loss, and optimal emission allocation. In the beginning, the exploration executed by Enhanced FDO is dependent on a simplistic fitness factor that delivers a less optimal solution by sticking into local minima and transforms some of its decision variables through their constraint violation. After applying the Sobol operator for population initialization and chaotic sin map for the optimal fitness, redistribution power operators are connected. The Enhanced operator ensures the feasibility of a probable solution that the thermal unit will take as an input and barely estimate the balance power constraint. Furthermore, the Enhanced population initialization approach consolidates a quasi-random Sabol sequence to create the initial solution in the multi-dimensional search space. A regular 24-unit system is applied with diverse power demands for experimental evaluation. Experiential results acquired utilizing the Enhanced variant of FDO confirm the superior performance in terms of low transmission loss, low fuel cost, and low emission allocation compared to the standard FDO. As a part of our future work, the authors are inspired by the hybridization of FDO with other metaheuristic algorithms such as BA, DE, and PSO. The authors aimed to take the best qualities from BA as local search capability, DE as optimal mutation factor, and PSO as inertia weight and incorporate them in FDO to achieve the best results. Furthermore, the authors are also interested in the fine-tuning of FDO parameters in combination with ELD constraint and their hyperparameter tuning. Additionally, the hybridized version of FDO will be evaluated to investigate the influence of objective evaluations on dimension reduction.

**Abbreviations**

Economic load dispatch (ELD), Prohibited Operating Zone (POZ), Valve-Point Effects (VPE), quantum bat algorithm (QBA), combined economic emission dispatch (CEED), evolution algorithm integrating with multiple mutation strategies (ADE-MMS), EVs and wind farms (WEV), Dynamic economic emission dispatching based on WEV system (WE_DEED), motion optimization algorithm (IMA), Bat Algorithm (BA), Artificial Bee Colony (ABC), Chaotic based Self-Adaptive (CSA), Learner Non-dominated Sorting Genetic Algorithm (NSGA-RL), Chaotic-crisscross differential evolution (CCDE), Differential Evolution algorithm (DEA), Self-adaptable differential evolution algorithm integrating with multiple mutation strategies (ADE-MMS), Modified crow search algorithm (MCSA), Multi-objective economic and environmental dispatch problem (EEDP), Coyote Optimization Algorithm (COA), Combined economic and emission dispatch (CEED), Emission-controlled economic dispatch (ECED), Fitness Dependent Optimizer (FDO), FDO with single-objective optimization problems (FDOSOOP), Multi-objective optimization problems (FDOMOOP).

**Funding:** The research received no funds.
**Conflict of Interest** The authors declare that they have no conflict of interest.
**Data Availability:** Data can be shared upon request from the corresponding author